\documentclass[letterpaper, 10 pt, conference]{ieeeconf}  

\usepackage{graphicx}
\usepackage{bm}

\IEEEoverridecommandlockouts                              

\usepackage{epsfig} 
\usepackage{mathptmx} 
\usepackage{times} 
\usepackage{amsmath} 
\usepackage{amssymb}  

\title{\LARGE \bf
Inspection-on-the-fly using Hybrid Physical Interaction Control for Aerial Manipulators
}

\author{Abbaraju Praveen$^{1}$,  Xin Ma$^{2}$, Harikrishnan Manoj$^{3}$, \\ Vishnunandan LN. Venkatesh$^{1}$, Mo Rastgaar$^{2}$, Richard M. Voyles$^{2}$ 

\thanks{Xin Ma is corresponding author. Email: maxin1988maxin@gmail.com.}%

\thanks{$^{1}$Abbaraju Praveen and Vishnunandan LN. Venkatesh  are PhD students in Purdue Polytechnic Institute, Purdue University, 47907, IN, USA {\tt\small \{pabbaraj, lvenkate\}@purdue.edu}}%

\thanks{$^{2}$Dr. Xin Ma is a post-doc and Drs. Mo Rastgaar and Richard Voyles are faculty in the School of Engineering Technology, Purdue Polytechnic Institute, Purdue University, 47907, IN, USA.
        {\tt\small \{ma633, rastgaar, rvoyles\}@purdue.edu}}%

\thanks{$^{3}$Harikrishnan M is an undergraduate research scholar in Purdue Polytechnic Institute, Purdue University, 47907, IN, USA. {\tt\small hmanojkr@purdue.edu}}%
}

\begin{document}

\maketitle
\thispagestyle{empty}
\pagestyle{empty}

\begin{abstract}

Inspection for structural properties (surface stiffness and coefficient of restitution) is crucial for understanding and performing aerial manipulations in unknown environments, with little to no prior knowledge on their state. Inspection-on-the-fly is the uncanny ability of humans to infer states during manipulation, reducing the necessity to perform inspection and manipulation separately. This paper presents an infrastructure for inspection-on-the-fly method for aerial manipulators using hybrid physical interaction control. With the proposed method, structural properties (surface stiffness and coefficient of restitution) can be estimated during physical interactions. A three-stage hybrid physical interaction control paradigm is presented to robustly approach, acquire and impart a desired force signature onto a surface. This is achieved by combining a hybrid force/motion controller with a model-based feed-forward impact control as intermediate phase. The proposed controller ensures a steady transition from unconstrained motion control to constrained force control, while reducing the lag associated with the force control phase. And an underlying Operational Space dynamic configuration manager permits complex, redundant vehicle/arm combinations. Experiments were carried out in a mock-up of a Dept. of Energy exhaust shaft, to show the effectiveness of the inspection-on-the-fly method to determine the structural properties of the target surface and the performance of the hybrid physical interaction controller in reducing the lag associated with force control phase.

\end{abstract}

\section{INTRODUCTION}

Recent advancements in multirotor Unmanned Aerial Vehicle (UAV) research, combining aerial mobility with manipulation has provided a platform for aerial manipulation tasks. UAVs are used for an array of physical interactive tasks including manipulation and grasping \cite{grasp} \cite{avian}, contact-based inspection \cite{manip} \cite{phyi}, and remote sensor placement \cite{boom} \cite{pushrelease}. Direct physical interaction with the environment provides the ability to perform an operation involving manipulation or inspection. Humans, through multiple modes of sensing are masters at understanding and adapting to the environment while interacting with it. For example, tactile sensing is used for various exploratory procedures to understand the surface/object properties before performing any manipulative procedures. In short, humans are experts at adapting to variations in the environment because they are constantly inspecting while interacting. We call this inspection-on-the-fly and feel it is a precursor to more intelligent and self-adaptive behaviors \cite{Cui2015} for all types of manipulative robots engaging in physical interactions.

\begin{figure}[thpb]
      \centering
      \includegraphics[width = 0.43\textwidth]{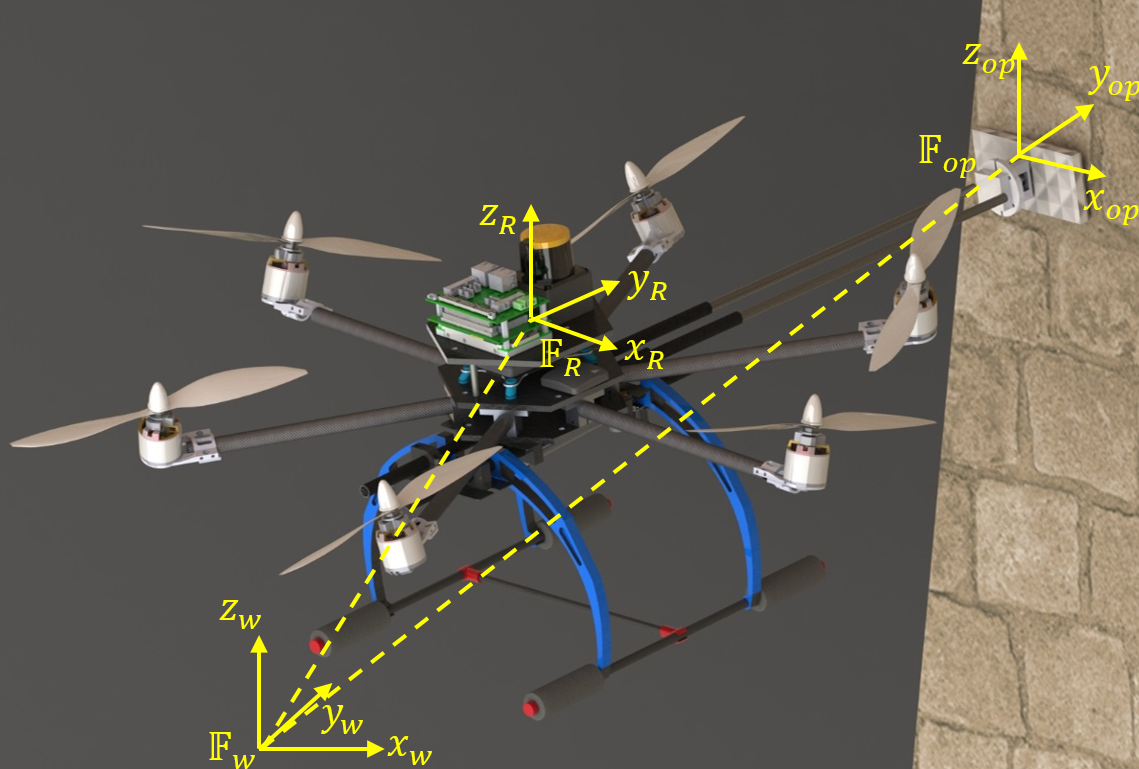}
      \caption{Structure of the fully-actuated UAV and the coordinate frames for
modeling \label{Dexhex}}
      \label{fig:DexHex1}
\end{figure} 

The challenges associated with unknown environment requires the robot to perform inspection-on-the-fly, to understand the physical state of the environment, while simultaneously performing interaction tasks. The unknown environments including remote and hostile locations, abandoned structures and nuclear facilities, which requires to employ robots as opposed to human workers. There is little to no prior knowledge about the physical state of the target surface and its accessibility. This raises the need to estimate and comprehend the states of such facilities before performing any meaningful operations. Some significant operations which require physical interaction are contact-based sampling of contaminated surfaces \cite{ssrr} \cite{WM}, applying chemical agents, surface drilling and, remote sensor placement \cite{boom} \cite{pushrelease}.

The current state-of-the art for physical inspections using aerial manipulators include various UAV platforms attached with robotic manipulator. While establishing a steady contact with the environment, the UAVs are prone to disturbances caused during interactions, eddy current wind disturbances, and any external disturbances. Quadrotors, despite of the limitations with under-actuation, have been employed for physical interaction tasks through motion control using flatness-based control \cite{flat}, adaptive sliding mode control and custom-arm mechanisms. However, a fully-actuated UAV with independent control over all 6D motions \cite{quadtilt} \cite{hextilt}, eliminates the problems associated with under-actuation and enables superior physical interaction capabilities.

Several physical interaction control strategies to enable steady interaction with the target surface while rejecting disturbances, are proposed in the literature using indirect force estimation and direct force feedback. A common strategy with indirect force estimation is admittance filter to control the interaction forces at the end-point of the rigid arm attached to a fully-actuated UAV \cite{6D}. \cite{port} has proposed a impedance controller derived by the energy-balancing passivity-based control technique for contact stability. However, the indirect force estimation using robot dynamics is subjected to uncertainties which doesn't correspond to just the interactions forces. Moreover, any external disturbance forces can't be distinguished from the interactions forces. However, with a direct force feedback, the interaction forces can be distinguished from external disturbance forces.

Early works in direct force feedback is proposed in \cite{impedance}, which uses the force sensor feedback in an impedance control loop for end-point compliance rather for interaction control. In \cite{direct}, a whole-body force control strategy for aerial manipulators is proposed for physical interaction control using hybrid position/force control. However, switching from unconstrained motion control to constrained force control has significant problems with impact forces \cite{paul}, which is not addressed in the aforementioned literature. The aerial manipulators are relatively less rigid and have lower coefficient of restitution, pushing them into inelastic collision leading to improper dissipation of energy during the impact. This results in a loss of contact or limit cycle behaviour, even under lower approach velocities. Therefore, we propose a hybrid physical interaction control for aerial manipulators based on energy dissipation, similar to the impact control techniques for fixed-based manipulators \cite{OPimpact}. A modified impact model based on the change in momentum with respect to time is used in the feed-forward impact controller.

The main contributions in this paper are a three-stage hybrid physical interaction controller and the infrastructure for the inspection-on-the-fly method being integrated into our ReFrESH self-adaptation architecture \cite{Cui2017}. Design and modelling of the aerial manipulator, along with the equations of motion, are presented in sec II. The hybrid physical interaction control, presented in sec. III, is aimed to provide steady transition from unconstrained motion to constrained force control using a model-based feed-forward impact controller. This controller ensures that the limit cycle behaviour and lag associated with switching directly from motion to force control are avoided. The ability to simultaneously perform explorations during  physical inspections is accomplished using the inspection-on-the-fly method. Sec. IV presents the models to estimate the coefficient of restitution (CoR) and surface stiffness which are significant for the aerial manipulator to understand and adapt to any variations present in the environment. Once the structural parameters/features are extracted they can be leveraged to enhance controller performance. Results of the experiments performed in a mock-up of a Dept. of Engergy exhaust shaft are presented in  sec. V, to show the effectiveness of the hybrid physical interaction controller in establishing a contact and determining the structural properties through inspection-on-the-fly method. The conclusion and future works are presented in sec. VI.

\section{MODELLING}

In this section, the structure and the model of a fully-actuated UAV equipped with a rigid arm are detailed. As shown in  Fig. 1, the UAV is modeled as a free-flyer base as in \cite{free}, which is actuated in the world frame, $\mathbb{F}_w (O_w: X_w, Y_w, Z_w)$. The state of the aerial manipulator is defined with respect to the world frame, $\mathbb{F}_w$ is arbitrarily placed such that the $z_w$ is opposite to gravity. The robot frame for the UAV, $\mathbb{F}_R (O_R:X_R, Y_R, Z_R)$ is placed with respect to the CoM of the UAV with the arm and the operational point fixed at the end-point, $\mathbb{F}_{op} (O_{op}:X_{op}, Y_{op}, Z_{op})$. The rigid arm is fixed to the UAV frame and is placed such that the $y_{op}$ is aligned with $y_R$ of the UAV at a distance, $d_{op} \in \mathbb{N}$. add variables to fig2 A spring loaded end-effector mechanism provides damping during interactions and helps to avoid structural damage. A 1D force sensor is placed at the end-effector to determine the force applied in $y_{op}$ direction. The UAV in the inertial frame is represented by $q_{R}= (x_{R}\ y_{R}\ z_{R}\ \theta_{R}\ \phi_{R}\ \psi_{R})^T \in \mathbb{R}^3 \times SO(3)$. The velocity w.r.t $q_{R}$ is given as, $\upsilon_R = J_R \upsilon_w$, where $J_R = [J_\upsilon J_\omega]^T \in \mathbb{R}^6$ is Jacobian mapping of velocities in $F_R$ to $F_w$, where $J_\upsilon \& J_\omega$ represent the Jacobian of the linear and angular velocities respectively. $\upsilon_R \in \mathbb{R}^3$ and $\upsilon_w \in \mathbb{R}^3$ are velocities in inertial and robot space respectively.

\subsection{Fully-actuated UAV Design}

As shown in Fig. 2, the structure of the fully-actuated hexrotor UAV, which is derived from a conventional hexrotor platform with tilted rotors, provides a set of thrust vectors that span the 6D Cartesian force space. The non-parallel actuator configuration is achieved by rotating adjacent rotors in opposite directions along the radial axis, referred to as \textit{cant} angle, $\alpha$. The critical UAV design parameters include the radius, $d$, and angle between each rotor, $\phi_i=60^0$. $c_f$ \& $c_q$ represent the rotor thrust constant and rotor torque constant. A non-zero \textit{cant} angle represents the non-parallel configuration and results in mapping the propeller angular speed, $\omega_i \in \mathbb{R}^i$ ($i=6$) to the force/torques at CoM, $\bm{F}_R \in \mathbb{R}^6$.

\begin{figure}[thpb]        
    \centering
    \includegraphics[width=0.9\linewidth,height=0.57\linewidth]{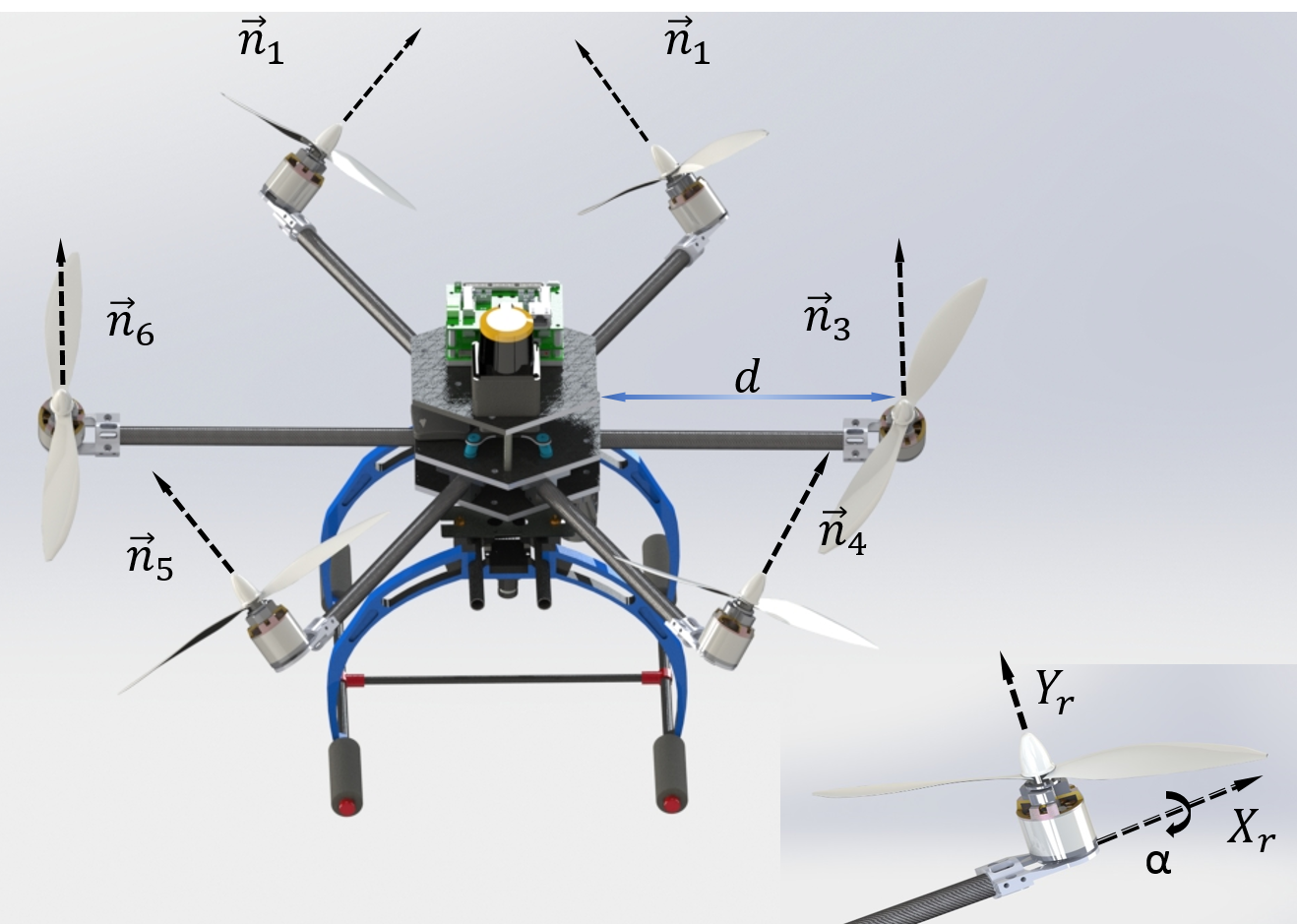}
    \label{fig:orient}
    \caption{Fully-actuated UAV design with non-parallel actuation mechanism.}
\end{figure}

\begin{equation}
    \textbf{F}_R = \begin{bmatrix} \bm{f}_R \\ \bm{\tau}_R \end{bmatrix} = M_{\alpha} \cdot \bm{\omega^2} \\ 
    \label{eq:mappin}
\end{equation}

where, $\bm{\omega}=[\omega_1 \ \omega_2 \ ... \ \omega_6]^T$ and $M_{\alpha} \in \mathbb{R}^{6 \times 6}$ with full rank.

\begin{equation}
    M_{\alpha} = \begin{bmatrix} c_fsin(\phi_i)sin(\alpha) \\ c_fcos(\phi_i)sin(\alpha) \\ c_fcos(\alpha) \\ sin(\phi_i)[c_qsin(\alpha)-dc_fcos(\alpha)] \\ cos(\phi_i)[c_qsin(\alpha)-dc_fcos(\alpha)] \\ (-1)^{i+1}[c_qcos(\alpha)+dc_fsin(\alpha)]  \end{bmatrix}
    \label{eq:mapmat}
\end{equation}

The choice of $\alpha$ angle is significant in determining the performance of the UAV in terms of arbitrary force exertion and lift forces. A common choice of optimization is to achieve uniformity in 6D force/torque exertion at hover. The optimization for \textit{cant} angle proposed in \cite{optimization} balances the maximum lateral forces and lift forces. Within the scope of this paper, a lateral force of approximately 2N is sufficient to maintain steady contact with a target surface. Therefore, the \textit{cant} angle is optimized with the objective functions, maximum lateral force of approximately 6N (to avoid actuator saturation) and maximum lift forces. This resulted in $\alpha = 25^\circ$ as the optimized \textit{cant} angle, which is used to conduct the experiments.

\subsection{Task Specifications} \label{sec:task}

For the scope of this paper, we assume the physical interactions are with a rigid and planar or non-planar surfaces. However, the proposed controller has the potential to extend to soft surfaces. At each of the control states in sec. III, a decoupled motion \& force control with respect to the direction of interaction is employed. Generalized state specification matrices, $\Omega$ \& $\Tilde{\Omega}$, in (\ref{eq:select}) are used to determine the unified operational command vector in $\mathbb{F}_{OP}$, for a decoupled control \cite{OP}. The task specification matrices, $\Sigma_f$ \& $\Sigma_\tau$ decouples the forces and torques with respect to task requirements. Here, $\Sigma = diag(\sigma_x,\sigma_y,\sigma_z)$ and the remainder directions are specified using $\overline{\Sigma}=I-\Sigma$. ($\sigma$ is binary)

\begin{equation}
    \Omega = \begin{bmatrix} S_f^T \Sigma_f S_f & 0 \\ 0 & S_\tau^T \Sigma_\tau S_\tau \end{bmatrix} ; \ \
    \Tilde{\Omega} = \begin{bmatrix} S_f^T \overline{\Sigma}_f S_f & 0 \\ 0 & S_\tau^T \overline{\Sigma}_\tau S_\tau \end{bmatrix} 
    \label{eq:select}
\end{equation}

In fig. (\ref{fig:DexHex1}), $y_{OP}$ is aligned with $y_r$ as the manipulator is fixed and the rotation transformation between them, $S_f$ \& $S_\tau$ become identity matrices. Task specifications are $\Sigma_f = diag(1,0,1)$ \& $\overline{\Sigma}_f = diag(0,1,0)$.

\subsection{Equations of Motion}

The force/torques in $\mathbb{F}_{OP}$ are computed w.r.t the jacobian transpose mapping in $\mathbb{F}_{W}$.

\begin{equation}
    \bm{F}_{op} = J^T_{op}(q_{op}) \cdot \bm{F}_W 
    \label{eq:OP}
\end{equation}

However, the forces in $\mathbb{F}_{OP}$ needs to be transformed to $\mathbb{F}_{R}$.  

\begin{equation}
    \bm{F}_{R} = \begin{bmatrix} I & \hat{\rho} \\ 0 & I \end{bmatrix} \cdot \bm{F}_{op} 
    \label{eq:OP1}
\end{equation}
where $\hat{\bm{\rho}}$ is the cross-product operator associated with the vector connecting the COM and the operational point.

\begin{equation}
    \bm{F}_{op} = J^T_{op}(q_{op}) \bm{\Lambda}(q_{op})\bm{F_{W}^*}+\bm{\widetilde{b_{r}}(q_{R}, \dot{q}_{R})}+\bm{g}
\end{equation}
where, $\Lambda(q_{op}) = [J_{op}(q_{op})\bm{A^{-1}(q_{R})}J^T_{op}(q_{op})]^{-1} \ \& \ A= diag \bm{(M\ M\ M\ I_{x}\ I_{y}\ I_{z})}$ and $M$ is the mass of the UAV and $I_{x},I_{y},I_{z}$ are the inertia. The desired motion command, $\bm{F_{W}}$, is given as

\begin{equation}
    \bm{F}_W = \bm{F}_W^m + \bm{F}_W^f + \bm{F}_W^{ccg}
\end{equation}

where, $\bm{F}_W^m, \ \bm{F}_W^f \ \& \ \bm{F}_W^{ccg}$ refer to the force commands for motion control, force control and Coriolis term \& centrifugal term, respectively. 

\section{HYBRID PHYSICAL INTERACTION}

Physical interaction requires the fully-actuated UAV to establish a steady contact with the target surface. Initial impact forces during the transition from unconstrained motion controlled approach to a constrained force controlled contact are controlled through a three stage hybrid control. The three states are defined as \textit{approach}, \textit{impact}, \textit{contact} and the transition between them is determined with a threshold detection in the force sensor measurement. In \textit{approach} state, a motion controller is used to move towards the target location until a large force threshold is detected. This indicates the initial contact and enables the \textit{impact} state. The impact energy is estimated from the approach velocity and dissipated during the \textit{impact} state through a feed-forward impulse model. This \textit{impact} state usually lasts on the order of tens of milliseconds before switching to \textit{contact} state. A direct force feedback controller manages steady contact in the \textit{contact} state. To exit the constrained force contact state at the end of task interaction, the \textit{approach} state is reinstated with a negative velocity impulse.



\subsection{State: \textit{Approach}} 

A velocity controller is employed in the direction normal to the target surface with position control for all other directions. The magnitude of the impact is directly proportional to the approach velocity, but the impact controller prevents bounce. Impact control requires estimates of the approach velocity, mass of the UAV, and CoR of the surface.

\begin{figure}[thpb]
    \centering
    \includegraphics[width=1\linewidth, height=0.2\linewidth]{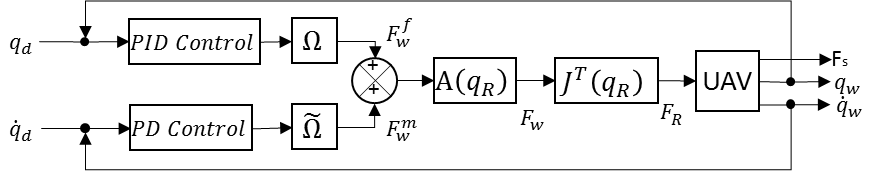}
    \caption{Velocity control on surface normal, and position control in the remainder directions.}
    \label{fig:approach}
\end{figure}

If the $y_{op}$ in Fig. \ref{Dexhex} is aligned normal to the target surface, then the velocity control is initiated in this direction. The reminder directions $x_{op} \ \& \ z_{op}$ are position controlled. Then the task specification matrices take the form as, $\Sigma_f = diag(1,0,1)$ \& $\overline{\Sigma}_f = diag(0,1,0)$. Fig. \ref{fig:approach} shows the control loop employed in the \textit{approach} state, where PID controller for position and PD controller for velocity.


\subsection{State: \textit{Impact}}

Impact forces are modelled as impulse force as they occur for a short period of time. Impulse force is defined as the change in momentum, $\rho$ 

\begin{equation}
    F_{impulse} = \Delta \rho/\Delta t
    \label{eq:momentum}
\end{equation}

The impact momentum is modelled as,
\begin{equation}
    F_w^{i}(t) = \int F_w^{i} \ dt = k_{vf} \cdot \Lambda(q_R) \cdot (v_f-v_i)
    \label{eq:impact}
\end{equation}

\begin{figure}[thpb]
    \centering
    \includegraphics[width=1\linewidth, height=0.2\linewidth]{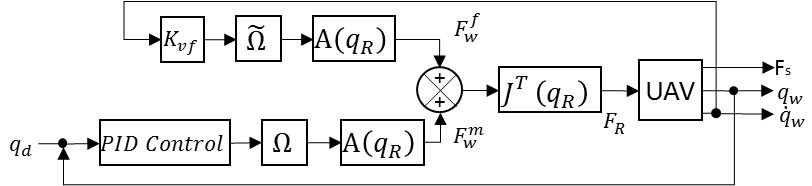}
    \caption{Model-based impact control on surface normal and position control in the remainder directions.}
    \label{fig:OPimp}
\end{figure}

Where, $K_{vf}$ gain determines the rate of energy dissipation during an impact, with respect to the change in velocity and is directly proportional to the CoR. Fig. \ref{fig:OPimp} shows the modified control loop for \textit{impact} state.

\subsection{State: \textit{Contact}}

\begin{figure}[thpb]
    \centering
    \includegraphics[width=1\linewidth, height=0.2\linewidth]{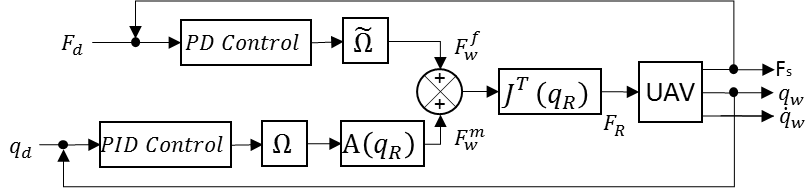}
    \caption{Force control on surface normal and position control in the remainder directions.}
    \label{fig:cont}
\end{figure}

After the impact forces are compensated by squelching the excess energy in the system, the direct force feedback control is activated in \textit{contact} state. Fig. \ref{fig:cont} shows the control loop to exert forces normal to the surface. 

However, additional trajectories for motion control can be applied to the reminder directions to accomplish required tasks. The constrained force control ensures no loss of contact throughout the task. 

\section{INSPECTION-ON-THE-FLY}

In this section, inspection-on-the-fly estimates structural properties of the target surface while performing the physical interaction task. The learning of structural properties is expected to be performed without the need for additional sensors or customized control algorithms. Structural stiffness and CoR are the two parameters that are significant for steady physical interactions. The energy lost during the \textit{impact} state is quantified using CoR, $e$, and is defined in (\ref{eq:restitution}) as the ratio of energy released during restitution to the energy stored during collision. 

\begin{equation}
    e = \frac{\dot{q}_{w_f}}{\dot{q}_{w_i}}
    \label{eq:restitution}
\end{equation}

\begin{figure}[thpb]
    \centering
    \includegraphics[width=0.6\linewidth, height=1.7cm]{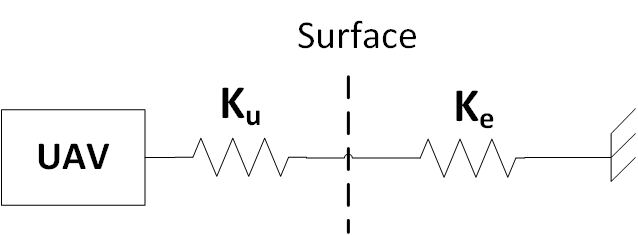}
    \caption{Model including stiffness of the mating structures}
    \label{fig:OPcontact}
\end{figure}

The surface stiffness significantly effects the force feedback interaction performance and is estimated by quantifying the impact model. Let's consider the stiffness of both the mating objects and model them with spring constants $K_u$, for UAV and $K_e$ for the target surface, usually fixed as shown in fig. \ref{fig:OPcontact}.

The stiffness of the target surface, $K_e$ is estimated using the model in (\ref{eq:spring}).
\begin{equation}
    K_e = (K_u \cdot K_t)/(K_u - K_t)
    \label{eq:spring}
\end{equation}

Where, $K_t$ is determined based on the total kinetic energy transition to potential energy during the initial impact phase,

\begin{equation}
    K_t = (0.5) \cdot m_{UAV} \cdot \dot{q}_{w_f}^2
    \label{eq:stiff}
\end{equation}

where, $\dot{q}_{w_f}$ is the velocity normal to the surface in world frame.

\section{EXPERIMENTS}

The experiments are conducted in a 9-foot diameter mock-up of the Waste Isolation Pilot Plant (WIPP) exhaust shaft (refer to video) using a fully-actuated hexrotor UAV attached with a rigid arm for swabbing the inside walls. The UAV is primarily controlled by an APM2.8 flight controller, which takes care of the low-level controls such as motor control and attitude stabilization. The choice of on-board sensors is based on the suitability for real-world applications. The sensor suite includes laser range finder fused with barometric pressure sensor data to estimate the altitude indoors, an optical flow sensor fused with IMU data estimates the velocity with respect to the ground plane. A 2D Lidar rangefinder will be able of provide the SLAM capabilities for motion in free space at a fixed altitude. The sensor processing is performed in Jetson TX2. A 1D force sensor at the tool tip will enable the force controlled physical interaction. The hybrid physical interaction control is implemented in a custom RTOS (real time operating system) along with the sensor fusion.

\begin{figure}[thpb]
    \centering
    \includegraphics[width=0.85\linewidth]{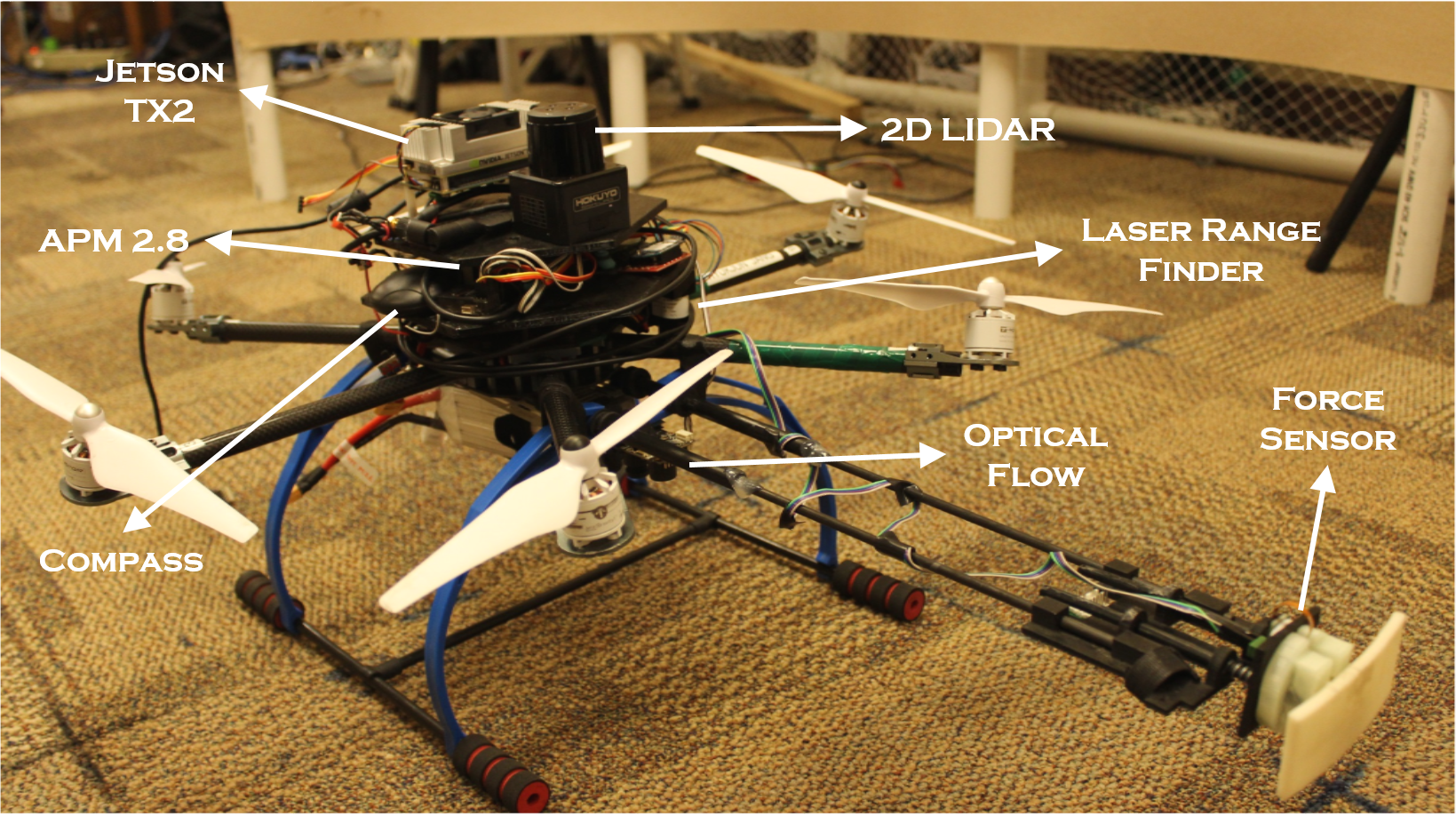}
    \caption{Fully-actuated UAV with the list of on-board sensors.}
    \label{fig:plots}
\end{figure}

\subsection{Hybrid Physical Interaction} 

\begin{figure}[thpb]
    \centering
    \includegraphics[width=1\linewidth]{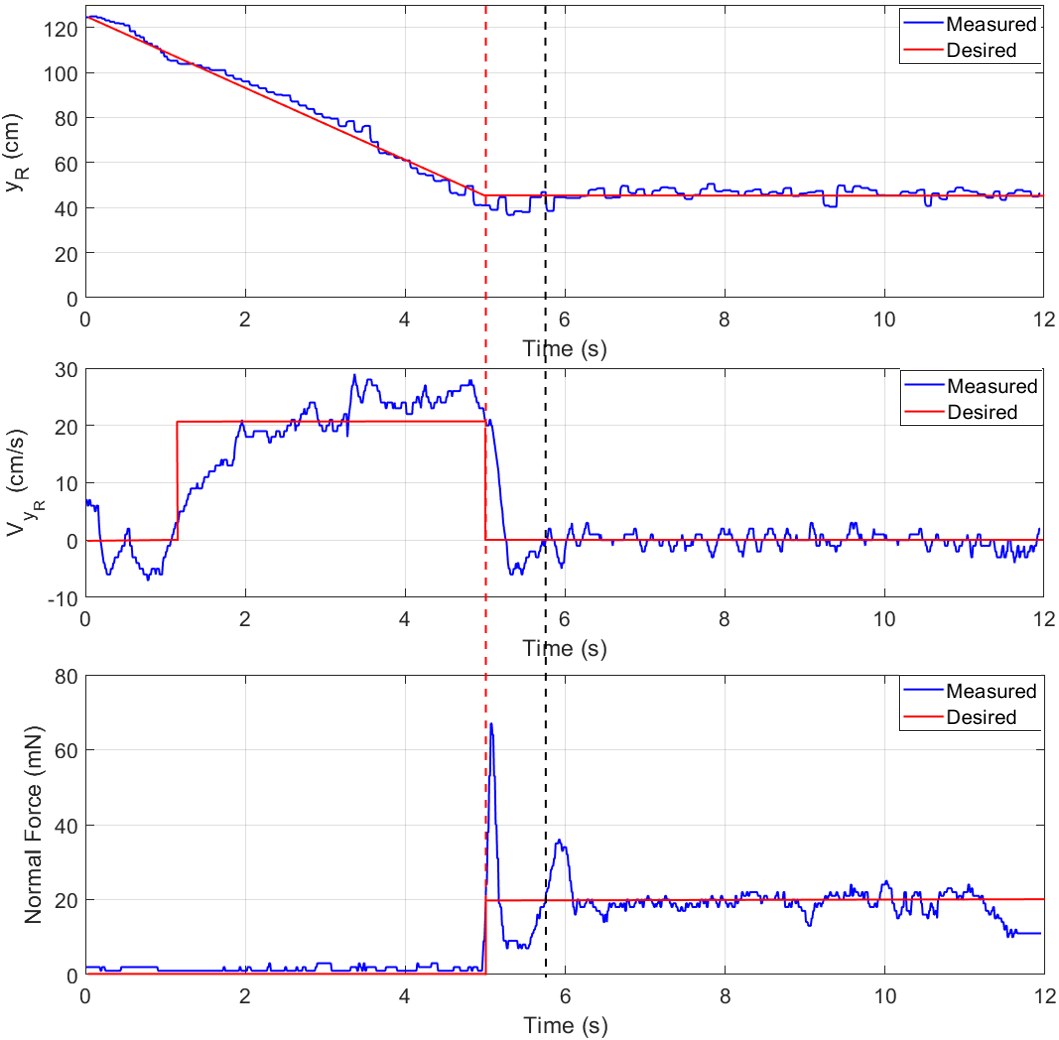}
    \caption{Three stage Hybrid physical interaction controller partitioned by the red and black dashed lines. Top: Measure position in $y_R$ with respect to the surface. Middle: Measured velocity with respect to UAV frame. Bottom: Measure force at the operational point.}
    \label{fig:plots_y}
\end{figure}

The performance of the proposed hybrid physical interaction control is evaluated based on the steady transition from unconstrained motion control to constrained force control without losing contact with the surface. Initially, the UAV is manually launched to a desired altitude and switched to autonomous mode. A trajectory controller positions the UAV at a nearby location close to the target surface. The hybrid physical interaction controller is enabled as soon as the UAV reaches the nearby location. The \textit{approach} state launches the velocity controller at 20cm/s as the set point as shown in the velocity plot of fig. \ref{fig:plots_y}. As the force sensors measures a non-zero value, the \textit{impact} state is enabled, to squelch the excess energy from the system. This impact phase is represented between the red and black dotted lines in fig. \ref{fig:plots_y}. 

\begin{figure}[thpb]
    \centering
    \includegraphics[width=1\linewidth]{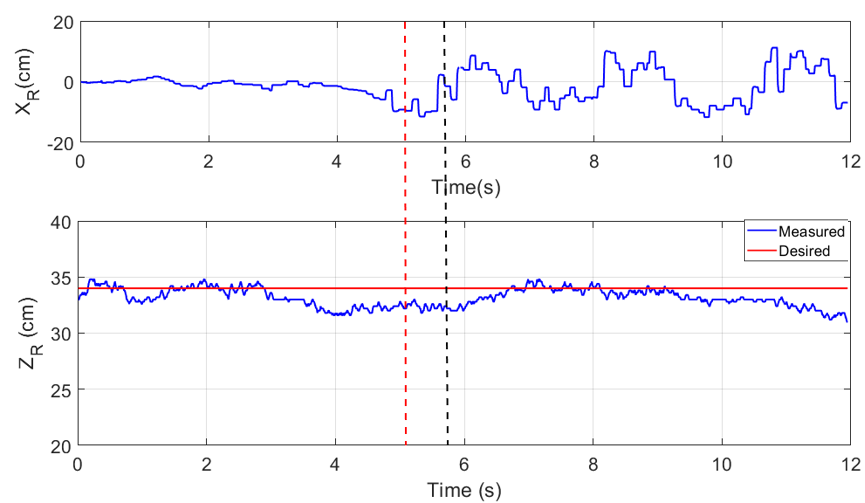}
    \caption{Top: Measure position in $x_R$ in the world frame. Bottom: Measured altitude with respect to ground.}
    \label{fig:plots_x}
\end{figure}

The impact controller ensures no loss of contact before switching to the force controller. The reduced errors in force, with an average error of $\pm 12mN$, are compensated in the force control phase. The contact is maintained for 6s before retrieving back to the initial location. The $x_R$ direction is position controlled throughout the experiment for which the measured position and velocity values are shown in fig.\ref{fig:plots_x}. Altitude during the operation is maintained at 34cm off the ground with an average error of $\pm 2cm$ as shown in fig. \ref{fig:plots_x}.

\subsection{Structural properties}

\begin{figure}[thpb] 
    \centering
    \includegraphics[width=0.85\linewidth, height=0.7\linewidth]{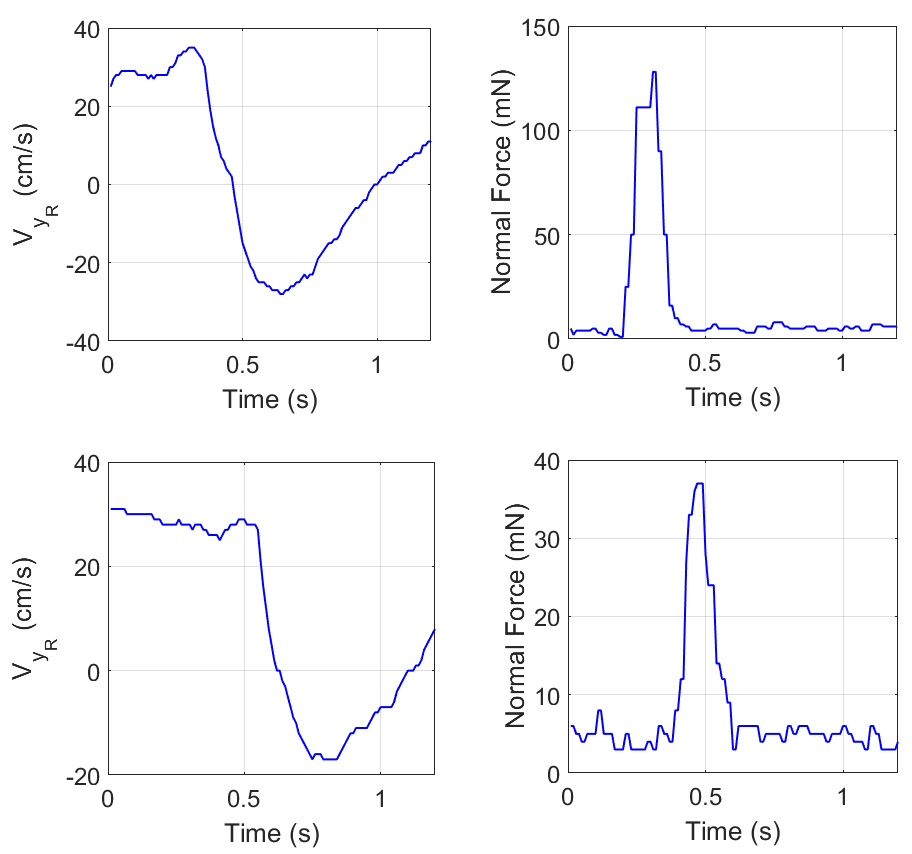}
    
\caption{Exploratory procedures to estimate the structural properties of the target surfaces, Top: surface S1 and Bottom: surface S2.}
\label{fig:properties}
\end{figure}

Inspection-on-the-fly method is used to estimate the structural properties of two different types of structures, wood (surface S1) and cardboard (surface S2). The goal of this experiment is to perform identify surface stiffness and CoR using the proposed hybrid physical interaction controller with low gains in the impact model. These low gains will reduce the significance of impact model and the UAV will bounce back to unconstrained motion. This phenomenon is showed in fig. \ref{fig:properties}, where the top plots are on surface S1 and the bottom plots are on surface S2. The measured approach velocity and final velocity along with the duration of contact are used to determine the surface stiffness and CoR, using (\ref{eq:spring}) \& (\ref{eq:restitution}) respectively. 

\begin{table}[h]
\caption{Estimated Structural Properties of Surfaces S1 \& S2}
\label{table:properties}
\begin{center}
\begin{tabular}{|c||c|c||c|c|}
\hline
\multicolumn{1}{|c||}{} & \multicolumn{2}{|c||}{Coefficient of Restitution} & \multicolumn{2}{|c|}{Surface Stiffness}\\
\hline
Iterations &Surface S2&Surface S1&Surface S2&Surface S1\\
\hline
 1 & 0.298 & 0.715 & 0.08 & 1.5\\
\hline
 2 & 0.326 & 0.823 & 0.11 & 2.3\\
\hline
 3 & 0.347 & 0.614 & 0.14 & 2.1\\
\hline
 4 & 0.381 & 0.689 & 0.17 & 1.9\\
\hline
\end{tabular}
\end{center}
\end{table}

This experiment is repeated 4 times, with different approach velocities and the estimated stiffness and CoR are presnted in table. \ref{table:properties}. The surface S2 has lower stiffness and CoR compared to S2, which shows that the surface S2 is softer compared to surface S1. This can also be observed in the velocity plots of fig. \ref{fig:properties}, where the negative velocity representing the bounce back, is higher on surface S1 compare to S2. These structural properties classifies that the surface S1 is harder surface compared to S2.

\section{CONCLUSIONS}

In this paper we have presented an inspection-on-the-fly method aimed at interacting with unknown surfaces using impact model based hybrid physical interaction control. Inspection-on-the-fly allows the aerial manipulator to infer states present in the environment while carrying out an active manipulation task. The results derived based on the coefficient of restitution and contact stiffness, the wood surface, S1 is classified to be harder than the cardboard surface, S2, resulting in a meaningful estimation of structural properties. During this procedure, the UAV used the hybrid physical interaction control to perform impact model based interaction with the target surface. The experiments conducted in mock-up exhaust shaft show the significance of controlling the impact phase in a physical interaction. 

Inspection-on-the-fly is a precursor in our approach as we deal with understanding a single state parameter for surfaces with multiple features/parameters. Current work in this paper shows how single state parameter estimation is sufficient to differentiate between materials at a fundamental level. The inspection-on-the-fly paradigm can be extended with added sensing/perception to estimate multiple parameters for a surface allowing a deeper understanding of the structures in the environment. A scope for future works would entail using tactile perception along with fusion of audio perception to detect textural properties for any surface.

\addtolength{\textheight}{-12cm}   




\section*{ACKNOWLEDGMENTS}

This work was supported, in part, by the Dept. of Energy, the NSF Center for Robots and Sensor for the Human Well-Being (RoSe-HUB) and by the National Science Foundation under grant CNS-1439717 and the USDA under grant 2018-67007-28439.



\begin{thebibliography}{99}

\bibitem{grasp} D. Mellinger, Q. Lindsey, M. Shomin, and V. Kumar, “Design, modeling, estimation and control for aerial grasping and manipulation,” in 2011 IEEE/RSJ International Conference on Intelligent Robots and Systems, Sept 2011, pp. 2668–2673.

\bibitem{avian} Thomas, Justin, Joe Polin, Koushil Sreenath, and Vijay Kumar. "Avian-Inspired Grasping for Quadrotor mini UAVs." ASME 2013 International Design Engineering Technical Conferences and Computers and Information in Engineering Conference: Volume 6A: 37th Mechanisms and Robotics Conference: V06AT07A014.

\bibitem{manip} F. Huber, K. Kondak, K. Krieger, D. Sommer, M. Schwarzbach, M. Laiacker, I. Kossyk, S. Parusel, S. Haddadin, and A. AlbuSchaffer, “First analysis and experiments in aerial manipulation using" fully actuated redundant robot arm,” in 2013 IEEE/RSJ International Conference on Intelligent Robots and Systems, Nov 2013, pp. 3452– 3457

\bibitem{phyi} M. Fumagalli, R. Naldi, A. Macchelli, F. Forte, A. Q. L. Keemink, S. Stramigioli, R. Carloni, and L. Marconi, “Developing an aerial manipulator prototype: Physical interaction with the environment,” IEEE Robotics Automation Mag., vol. 21, no. 3, pp. 41–50, Sept 2014.

\bibitem{boom} D. McArthur, A. Chowdhury, D. Cappelleri. "Design of the I-BoomCopter UAV for Environmental Interaction", IEEE International Conference on Robotics and Automation (ICRA). Marina Bay Sands, Singapore, May 29 - June 3, 2017.

\bibitem{pushrelease} S. Hamaza et al., "Sensor Installation and Retrieval Operations Using an Unmanned Aerial Manipulator," in IEEE Robotics and Automation Letters, vol. 4, no. 3, pp. 2793-2800, July 2019.

\bibitem{ssrr} G. Jiang and R. Voyles, “Hexrotor UAV platform enabling dexterous interaction with structures-flight test,” in 2013 IEEE International Symposium on Safety, Security, and Rescue Robotics (SSRR), Oct 2013, pp. 1--6.

\bibitem{WM} P. Abbaraju, R. Voyles, "Sensing and Sampling of Trace Contaminations by a Dexterous Hexrotor UAV at Nuclear Facilities-18600", Proc. of WM2018 Symposium, March 18--22, Phoenix, Arizona, USA.

\bibitem{flat} S. Formentin and M. Lovera, "Flatness-based control of a quadrotor helicopter via feedforward linearization," 2011 50th IEEE Conference on Decision and Control and European Control Conference, Orlando, FL, 2011, pp. 6171-6176.

\bibitem{quadtilt} D. Brescianini and R. D’Andrea, “Design, modeling and control of an omni-directional aerial vehicle,” in 2016 IEEE International Conference on Robotics and Automation (ICRA), May 2016, pp. 3261– 3266.

\bibitem{hextilt} M. Ryll, H. H. Bulthoff, and P. R. Giordano, “Modeling and Control of a Quadrotor UAV with Tilting Propellers,” in Robotics and Automation (ICRA), 2012 IEEE Intl Conf on, May 2012, pp. 4606– 4613.

\bibitem{6D} Ryll, M., Muscio, G., Pierri, F., Cataldi, E., Antonelli, G., Caccavale, F., … Franchi, A. (2019). 6D Interaction Control with Aerial Robots: The Flying End-Effector Paradigm. The International Journal of Robotics Research, 38(9), 1045–1062.

\bibitem{Cui2015} Y. Cui, R.M. Voyles, J.T. Lane, A. Krishnamoorthy, M.H. Mahoor, "A Mechanism for Real-Time Decision Making and System Maintenance for Resource Constrained Robotic Systems through ReFrESH," in Autonomous Robots 39 (4), 487-502, 2015.

\bibitem{Cui2017} Y. Cui, R.M. Voyles, X. Zhao, J. Bao, E. Bond, "A Software Architecture Supporting Self-Adaptation of Wireless Control Networks," in IEEE Intl Conf on Automation Science and Engineering (CASE), pp. 346-351, 2017.

\bibitem{port} R. Rashad, F. Califano, and S. Stramigioli, “Port-hamiltonian passivity-based control on se (3) of a fully actuated UAV for aerial physical interaction near-hovering,” IEEE Robotics and Automation Letters, vol. 4, no. 4, pp. 4378–4385, 2019.

\bibitem{impedance} G. Antonelli, E. Cataldi, G. Muscio, M. Trujillo, Y. Rodriguez, F. Pierri, F. Caccavale, A. Viguria, S. Chiaverini, and A. Ollero, “Impedance control of an aerial-manipulator: Preliminary results,” in 2016 IEEE/RSJ Int. Conf. on Intelligent Robots and Systems, Daejeon, South Korea, 2016, pp. 3848–3853.

\bibitem{direct} G. Nava, Q. Sablé, M. Tognon, D. Pucci and A. Franchi, "Direct Force Feedback Control and Online Multi-Task Optimization for Aerial Manipulators," in IEEE Robotics and Automation Letters, vol. 5, no. 2, pp. 331-338, April 2020.

\bibitem{paul} R. Paul, "Problems and research issues associated with the hybrid control of force and displacement," Proc of IEEE Intl Conf on Robotics and Automation, Raleigh, NC, USA, 1987, pp. 1966-1971.

\bibitem{impac} T. Bartelds, A. Capra, S. Hamaza, S. Stramigioli and M. Fumagalli, "Compliant Aerial Manipulators: Toward a New Generation of Aerial Robotic Workers," in IEEE Robotics and Automation Letters, vol. 1, no. 1, pp. 477-483, Jan. 2016.

\bibitem{OPimpact} O. Khatib and J. Burdick, "Motion and force control of robot manipulators," Proceedings. 1986 IEEE Intl Conf on Robotics and Automation, San Francisco, CA, USA, 1986, pp. 1381-1386.

\bibitem{optimization} G. Jiang, R. Voyles, K. Sebesta and H. Greiner, "Estimation and optimization of fully-actuated multirotor platform with nonparallel actuation mechanism," 2017 IEEE/RSJ Intl Conf on Intelligent Robots and Systems (IROS), Vancouver, BC, 2017, pp. 6843--6848.

\bibitem{OP} O. Khatib, "A unified approach for motion and force control of robot manipulators: The operational space formulation," IEEE Journal on Robotics and Automation, vol. 3, no. 1, pp. 43--53, February 1987.

\bibitem{free} J. Russakow, O. Khatib and S. M. Rock, "Extended Operational Space Formulation for Serial-to-Parallel Chain (branching) Manipulators," Proceedings of 1995 IEEE Intl Conf on Robotics and Automation, Nagoya, Japan, 1995, pp. 1056-1061 vol.1.

\end{thebibliography}
\end{document}